# A Computational Model of Afterimages based on Simultaneous and Successive Contrasts

Jinhui Yu[1] & Kailin Wu[1], Kang Zhang[2], Xianjun Sam Zheng[3]

[1]State Key Lab of CAD&CG, Zhejiang University

[2] Department of Computer Science, University of Texas at Dallas

[3]Siemens Corporate Research, Inc

**Abstract** Negative afterimage appears in our vision when we shift our gaze from an over stimulated original image to a new area with a uniform color. The colors of negative afterimages differ from the old stimulating colors in the original image when the color in the new area is either neutral or chromatic. The interaction between stimulating colors in the test and inducing field in the original image changes our color perception due to simultaneous contrast, and the interaction between changed colors perceived in the previously-viewed field and the color in the currently-viewed field also affects our perception of colors in negative afterimages due to successive contrast. Based on these observations we propose a computational model to estimate colors of negative afterimages in more general cases where the original stimulating color in the test field is chromatic, and the original stimulating color in the inducing field and the new stimulating color can be either neutral or chromatic. We validate our model with human experiments.

## 1. Introduction

When we stare at a color image, say a circular test field surrounded by a large rectangular inducing field, for several seconds and then shift our focus to a new area with a uniform color, we can see a negative afterimage (afterimage for short) with the same shape as the original image but different colors. The same effect will occur while the center of gaze is fixed, the stimulus changes temporally from old stimulating colors to the new stimulating color (in the experimental environment). The afterimage is believed to arise from (i) bleaching of photochemical pigments [1,2,3,4,5,6], or (ii) neural adaptation [7,8,9,10,11,12,13,14,16,17,18,19,20]. However, photochemical pigment bleaching hypothesis fails to explain the afterimage we see with closed eyes in a dark room, because it does not make clear what generates this long-lasting signals that make fixed images inside our closed eyes without external photonic source. The recent hypothesis of delayed bioluminescent photons [21] attempts to explain the afterimage seen with eyes closed in the dark room.

As for colors of afterimages, most of previous studies state that the color relationships between the original image and afterimage are complementary. However two groups of complementary color pairs are mentioned in the literature. The first group of color pairs are red- green, yellow-purple and blue-orange, that is, if the original image was red, the afterimage will be green, or if the image was green, the afterimage will be red [8,22,23]. The second group of color pairs are red–cyan (green-blue), green–magenta (blue-red), and blue–yellow, as reported in [24,25,26] that staring at red generates a cyan afterimage, staring at green generates a magenta afterimage, or staring at blue generates a yellow afterimage. Although some researchers [21] attribute these subjective differences in afterimages to the chromatic adaptation [27], we find that the color relationship between the original image and afterimage mentioned in the second group of complementary pairs is more consistent with our subjective experiences, and this is also confirmed by a recent study by Manzotti [28]. To avoid possible confusion, we use the term *opposite colors* in RGB color model to describe color pairs mentioned in the second group. When those opposite color pairs are placed next to each other, they create the strongest contrast.

In contrast to colors, strengths of afterimage are seldom addressed in the literature. Chromatic adaptation results changes in hue, saturation and brightness [27], and this may partly explain variations of afterimages in strength, and a distinct property of afterimages is their increased blurring overtime [29]. Studies [30] have shown that effect using sinusoidal gratings of different colors, orientation and frequency as stimuli on the quantitative measurements of blurriness. The presence of glare does not trigger any strengthening of afterimages [31] and this provides further evidence for a retinal explanation of afterimage.

Afterimages have been studied extensively in the cases where the new stimulating color is neutral (white or gray). However, afterimages also appear in the area with a chromatic color, and, in fact, colors of afterimages appearing in the latter cases (chromatic color) differ dramatically from those appearing in the former ones (neutral color).

Manzotti [28] makes a different hypothesis, in which afterimages are regarded as partial and temporary blindness to color components. Specifically, if S is the stimulating color in the original image, A is the color of the afterimage and B is the stimulating color in the new background, a simple relation can be devised: A = B – k S, where k is a modulating parameter. This equation may correctly predict afterimage colors when (B-kS)>0. For instance, S is red, B is white, and A is white minus red, which is cyan: A = white – red = cyan [28]. But the author fails to offer any predictions if A is negative when (B-kS)<0.

Moreover, the interaction between two colors in the original image has been largely ignored in the previous studies of afterimages. This interaction is called as *simultaneous contrast*, which generally refers to the tendency for a sensation such as lightness, color, or warmth to induce the opposite sensation in a stimulus with which it is juxtaposed [32]. Another visual effect related to simultaneous contrast is *successive contrast*, which refers to the phenomenon whereby a sensation such as lightness, color, or warmth tends to induce the opposite sensation in a stimulus that follows it [32]. Most of previous studies of afterimages focus primarily on the opposite colors inducing in afterimages, and neglect largely the role of the new stimulating color played in afterimages.

It is interesting to note that the process of afterimage production corresponds to the simultaneous contrast and successive contrast in succession. These observations lead us to investigate the color relationship between the original image and afterimage in more general cases where the old stimulating color in the test field is chromatic, and the old stimulating color in the inducing field and the new stimulating color can be either neutral or chromatic. Our goal is to computationally estimate the colors of afterimage, rather than to investigate how afterimages are produced in these conditions. Our model provides useful insights to what mostly influence afterimage colors. Based on our experiments, we develop a model to predict colors of afterimages accurately when the old and new stimulating colors change among distinctive colors (red, green, blue, cyan, magenta, yellow, white and black) in both test and inducing fields..

## 2. Estimation of afterimage colors

Previous studies compared the effects of using adjacent or nearby inducing fields or backgrounds that completely surround the test field, and found the greatest effects on perception for small test fields with large inducing surrounds [33,34]. We also confine our study to the cases where the test field is a circular area surrounded by a large rectangular inducing field (as indicated in Fig. 1). Based on this setting, we estimate afterimage colors in the test and inducing fields in slightly different fashions. Before describing our model in detail, we introduce our notation.

**Notation.** We adopt the RGB color model and denote a color C with its three components as C(R,G,B), with values assigned to each component are normalized. With this notation, multiplying the color C with a coefficient α can be written as $\alpha C=C(\alpha R, \alpha G, \alpha B)$, the opposite color C' of C as $C'=1-C=C'(1-R,1-G,1-B)$, and mixing two colors $C_1$ and $C_2$ to obtain a new color $C_3$ as $C_3=\alpha C_1+(1-\alpha)C_2$, where $\alpha<1$ is the weight.

Referring to Fig.1(a) and (b), the original stimulating color in the test field and inducing field are denoted by $C_{OT}(R_{OT}, G_{OT}, B_{OT})$ and $C_{OI}(R_{OI}, G_{OI}, B_{OI})$, respectively, the new stimulating color by $C_N(R_N, G_N, B_N)$, where the subscript O stands for original, T and I for test and inducing, N for new.

The interaction between $C_{OT}$ in the test field and $C_{OI}$ in the inducing field together at the simultaneous contrast stage results in a modified color $C_{MT}$ in the test field (Fig. 1(c)), and the interaction between $C_{MT}$ in the test field and the new stimulating color $C_N$ at the simultaneous contrast stage produces afterimage color $C_{AT}$ in the test field (Fig. 1(d)). In the estimation of the color $C_{AI}$ in the inducing field of afterimages, only the interaction between $C_{OI}$ in the inducing field and the new stimulating color $C_N$ due to successive contrast is taken into account.

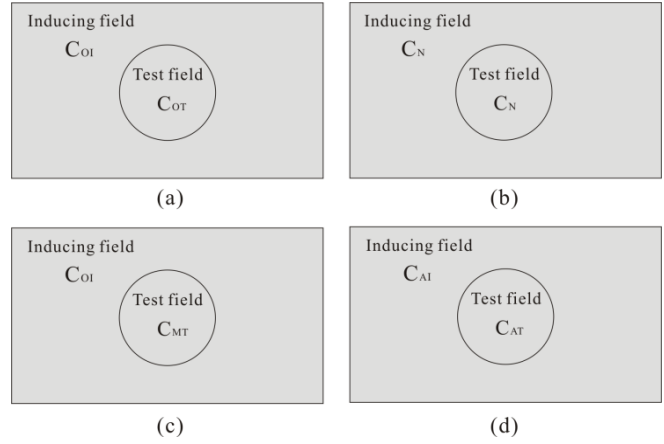

Fig. 1. Illustrations of notations: (a) and (b): physical stimulating colors; (c) and (d): subjective colors due to simultaneous contrast and successive contrast.

We propose a model consisting of three parts, (1) estimate of modified color in the test field in the original image, (2) opposite colors induced in the test and inducing fields in the afterimage, and (3) estimate of colors in the test and inducing fields in the afterimage.

**Part 1:Estimation of modified color in the test field in the original image.** When the color $C_{OI}$ in the inducing field is chromic, its opposite color $C'_{OI}$ can be obtained by $C'_{OI}(1-R_{OI},1-G_{OI},1-B_{OI})$. The modified color in the test field, denoted by $C_{MT}(R_{MT},G_{MT},B_{MT})$ where M stands for modified, can be calculated by mixing $C'_{OI}$ with $C_{OT}$, that is, $C_{MT}=\alpha C'_{OI}+(1-\alpha)C_{OT}$, where α is set at 0.4 experimentally.

White color is extensively used as surrounds in the studies of afterimages. In this case, the color $C_{OT}$ in the test field appears darker than pure white due to simultaneous contrast. Since the opposite color of white is black, the modified color in the test field $C_{MT}$ can be obtained simply by $C_{MT}=(1-\alpha)C_{OT}$, because all RGB components in $C'_{OI}$ are all zeros.

**Part 2:Opposite colors induced in the test and inducing fields in the afterimage.** The opposite color $C'_{MT}$ of the modified color $C_{MT}$ is induced in the test field of the afterimage, and can be obtained by $C'_{MT}(1-R_{MT},1-G_{MT},1-B_{MT})$. Similarly, the opposite color $C'_{OI}$ of the old stimulating color $C_{OI}$ is induced in the afterimage's inducing field, and can be obtained by $C'_{OI}(1-R_{OI},1-G_{OI},1-B_{OI})$.

**Part 3: Estimate of colors in the test and inducing fields in the afterimage.** When our eyes move to the different area with the new stimulating color $C_N$, the successive contrast stage is triggered. We mix $C'_{MT}$ with $C_N$ to estimate the color $C_{AT}$ in the afterimage's test field, $C_{AT}=\beta_T C'_{MT}+(1-\beta_T)C_N$, where $\beta_T$ is set to 0.4 through experiments. Similarly, the color in the afterimage's inducing field is estimated by $C_{AI}=\beta_I C'_{OI}+(1-\beta_I)C_N$, where $\beta_I$ is set to 0.1 when $C_{OI}$ is white, or 0.2 when $C_{OI}$ is chromatic, through experiments. We will discuss different settings of $\beta_I$ in the next section.

## 3. Formulation of our model

The modified color $C_{MT}$ in the test field can be expressed as the weighted sum of the opposite color $C'_{OI}$ of the stimulating color $C_{OI}$ in the inducing field and the stimulating color $C_{OT}$ in the test field in the original image:

$$C_{MT}=\alpha C'_{OI}+(1-\alpha)C_{OT} \quad (1)$$

The estimated color $C_{AT}$ in the test field of afterimages can be expressed as the weighted sum of the opposite color $C'_{MT}$ of the modified color $C_{MT}$ in the test field in the original image and the new stimulating color $C_N$,

$$C_{AT}=\beta_T C'_{MT}+(1-\beta_T)C_N \quad (2)$$

By substituting $C'_{OI}$ with $(1-C_{OI})$ in (1) and $C'_{MT}$ with $(1-C_{MT})$ in (2), we can obtain the following formula:

$$C_{AT}=\beta_T(1-\alpha)(1-C_{OT})+\beta_T\alpha C_{OI}+(1-\beta_T)C_N \quad (3)$$

It is interesting to note from Formula (3) that the first term corresponds to the opposite color of the old stimulating color $C_{OT}$ in the test field, the second term corresponds to the old stimulating color $C_{OI}$ in the inducing field, and the third term relates to the new stimulating color $C_N$. It is somewhat odd to note that $C_{OI}$ in the inducing field also plays a role on the color in the test field of afterimages. Since the opposite color $C'_{OI}$ of $C_{OI}$ is used to calculate the modified color $C_{MT}$ in the test field at the simultaneous contrast stage, and $C'_{OI}$ is further used to calculate the opposite color $C'_{MT}$ of $C_{MT}$ at the successive contrast stage, as a result, the old stimulating color $C_{OI}$ in the inducing field finally appears in Formula (3) with a positive coefficient $\beta_T\alpha$.

Let $\beta_T=0.4$ and $\alpha=0.4$ in (3), as set in Section 2 for the cases where $C_{OT} \neq C_N$, we have the following formula for the color $C_{AT}$ in the test field of afterimages:

$$C_{AT}=0.24(1-C_{OT})+0.16C_{OI}+0.6C_N \quad (\text{when } C_{OT} \neq C_N) \quad (4)$$

Formula (4) shows that the opposite color of the old stimulating color $C_{OT}$ in the test field has a weight of 24% in the test field of afterimages, the old stimulating color $C_{OI}$ in the inducing field has 16% weight, and the new stimulating color $C_N$ has 60% weight.

Compared with the color in the test field of afterimages, the formula for estimating the color $C_{AI}$ in the inducing field of afterimages is simple:

$$C_{AI}=\beta_I(1-C_{OI})+(1-\beta_I)C_N \quad (5)$$

Furthermore, when $C_{OI}$ is white, the opposite color of $C_{OI}$ in the inducing field is black, thus only the second term $(1-\beta_I)C_N$ contributes to $C_{AI}$. Since the afterimage look somewhat duller compared with the original image, we set $\beta_I$ to be 0.1 in this case. When $C_{OI}$ is chromatic (other than white), the first term in Formula (5) is then no longer zero, and we found from experiments that $C_{AI}$ is consistent with that in our vision when $\beta_I=0.2$.

**4. Parameter setting for special cases**

The model in (4) is able to predict colors in the test field of afterimages accurately as shown later when old stimulating colors in the test and inducing fields in the original image and new stimulating color vary among distinctive colors such as red, green, blue, cyan, magenta, yellow, white and black. We also notice that afterimage colors predicted by this model are somewhat less satisfactory in a few special cases, in particular when the inducing field is kept white in the original image, the new stimulating color $C_N$ is the same with the old stimulating color $C_{OT}$ in the test field in the original image. These cases are rarely addressed in the previous studies of afterimages. In order to get some insights about how RGB components of $C_{AT}$ vary when $C_{OI}$ is white and $C_N=C_{OT}$, we rewrite the formula in (3) with their RGB components of $C_{AT}$, $C_{OT}$ and $C_N$:

$$R_{AT}=\beta_T(1-\alpha)(1-R_{OT})+\beta_T\alpha R_{OI}+(1-\beta_T)R_N$$
$$G_{AT}=\beta_T(1-\alpha)(1-G_{OT})+\beta_T\alpha G_{OI}+(1-\beta_T)G_N \quad (6)$$
$$B_{AT}=\beta_T(1-\alpha)(1-B_{OT})+\beta_T\alpha B_{OI}+(1-\beta_T)B_N$$

When $C_{OT}$ and $C_N$ are red, three formulae in (6) become: $R_{AT}=(1-\beta_T)R_N+\beta_T\alpha R_{OI}$, $G_{AT}=\beta_T(1-\alpha)+\beta_T\alpha G_{OI}$, $B_{AT}=\beta_T(1-\alpha)+\beta_T\alpha B_{OI}$. Here the second term remains the same in $R_{AT}$, $G_{AT}$ and $B_{AT}$, the first term, which is a constant, remains the same in $G_{AT}$ and $B_{AT}$, thus only the first term in $R_{AT}$ plays a differential role in the afterimage. We know from the spectral sensitivity [35] in color vision that red is least sensitive among red, green and blue, the coefficient $(1-\beta_T)$ of $R_N$ should be therefore set higher. Also, the formula in (1) can be rewritten with their RGB components of $C_{OI}$ and $C_{OT}$:

$$R_{MT}=\alpha R'_{OI}+(1-\alpha)R_{OT}$$
$$G_{MT}=\alpha G'_{OI}+(1-\alpha)G_{OT} \quad (7)$$

$$B_{MT}=\alpha B'_{OI}+(1-\alpha)B_{OT}$$

We notice that when $C_{OI}$ is white, its opposite color $C'_{OI}$ is black, thus only the second term in (1) is left. Further, when the color $C_{OT}$ in the test field is red, the coefficient $(1-\alpha)$ of $R_{OT}$ in (7) should be set higher due to spectral sensitivity.

Similarly, when $C_{OT}$ and $C_N$ are green, three formulae in (6) become $R_{AT}=\beta_T(1-\alpha)+\beta_T\alpha G_{OI}$, $G_{AT}=(1-\beta_T)G_N+\beta_T\alpha R_{OI}$, $B_{AT}=\beta_T(1-\alpha)+\beta_T\alpha B_{OI}$, the coefficient $(1-\beta_T)$ of $G_N$ in $G_{AT}$ should be set lower because green is most sensitive among red, green and blue, also the coefficient $(1-\alpha)$ of $G_{OT}$ in (7) should be set lower. Through experiments guided by the spectral sensitivity in color vision, we finally set $\alpha=0.6$, $\beta_T=0.35$ when $C_{OT}$ and $C_N$ are red; $\alpha=0.7$, $\beta_T=0.4$ when $C_{OT}$ and $C_N$ are blue, $\alpha=0.75$, $\beta_T=0.45$ when $C_{OT}$ and $C_N$ are green.

## 5. Verification of the model

In order to compare afterimages predicted by our model with previous complementary afterimages, we design a user interface (UI) shown in Fig. 2(a). The circular test field with the old stimulating color $C_{OT}$ and rectangle inducing field with the old stimulating color $C_{OI}$ are displayed on the upper window. When a user clicks the "start" button and is asked to stare at the center of the circle for 20s, our UI draws entire upper window with the new stimulating color $C_N$, and an afterimage will appear in the user's vision. Also, two smaller images denoted by S1 and S2 are displayed in two bottom windows (initially set to be gray) immediately after the new stimulating color replaces old ones in the upper window. In S1, the test field with the complimentary color of $C_{OT}$ is drawn over the inducing field with color $C_N$, the brightness of S1 is lowered by multiplying S1 with 0.9. In S2, the test field and inducing field are drawn with their estimated colors of afterimages by our model. To simulate the fuzziness and leaky edges of the shape in the afterimage [36], we apply Gaussians smooth filters to afterimages in S1 and S2. Also, S1 and S2 are randomly displayed either left or right at the bottom of the window during each test.

Users click the area of S1 or S2 to pick an image that they think better matches with the afterimage that they see. If a user cannot make a choice, s/he may click "redo" button to repeat the same test. When the user is confident about his or her choice, , s/he then clicks the "finish" button to end this trial , and our UI takes a record by assigning score 1 to the confirmed image. If the user has difficulty to judge which image is closer to the afterimage in their vision, s/he may click "Almost the same" button and our UI assigns 0.5 to S1 and S2 respectively.

In our tests we keep the old stimulating color white denoted by WHITE(1,1,1) in the inducing field. The old stimulating colors in the test field varies from red RED(1,0,0), green GREEN(0,1,0) to blue BLUE(0,0,1). The new stimulating colors vary from white, black BLACK(0,0,0), red, green and blue. Thus there are 15 tests in total. We asked 15 users with normal color vision (Male 8, female 7, averaged age 22) to conduct tests, and results of the experiment are reported next.

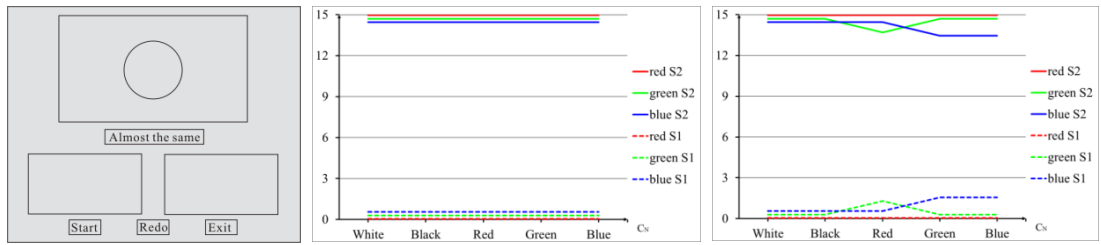

Fig. 2. (a): Interface of our tests; (b) and (c): Scores for smaller images S1 and S2 picked up by users.

## 6. Test results and comparisons

In total we conducted two groups of tests for comparing afterimages predicted by our model with previous complimentary afterimages using two groups of color pairs mentioned in Section 1, respectively. Corresponding results of two groups of tests are shown in Fig. 2(c) and (d), where scores assigned to S1 and S2 are plotted for the circular shape with different old stimulating colors indicated by red, green and blue lines (we draw three color lines slightly apart vertically to avoid overlaps between different color lines caused by the same scores assigned to them), the horizontal axis corresponds to the new stimulating color $C_N$, and vertical axis corresponds to scores assigned to S1 and S2 during tests. Solid lines correspond to the afterimage S2 predicted by our model and dashed lines correspond to previous complementary afterimages S1.

Statistics in Fig. 2(b) show that complementary afterimages obtained with color pairs described in the first group of color relationships (red-green, yellow-purple and blue-orange) visually are so different from those in their vision, thus the users selected afterimages predicted by our model without any doubt, as a result, all 15 scores are assigned to the afterimages predicted by our model.

Statistics in Fig. 2(c) shows that complementary afterimages obtained with color pairs described in the second group of color relationships (red–cyan, green–magenta, and blue–yellow) visually are also different from those in the user's vision in general, with slight variations in scores assigned to complementary afterimages and those predicted by our model over different new stimulating colors. When the new stimulating color is white and black, afterimages predicted by our model are clearly better than the complementary afterimages, because all 15 users picked up afterimages predicted by our model. When new stimulating color is changed into red, green and blue, afterimages predicted by our model are also much better than complementary afterimages, because more than 13 out of 15 scores are assigned to afterimages predicted by our model.

## 7.  More results

In subsequent Fig. 3~6, we show afterimages in more general cases where the old stimulating colors in the test and inducing fields and the new stimulating color vary more dynamically. There are 4 subfigures in those figures, (a) is a circular test field inside a rectangular inducing field with their old stimulating colors, (b) shows the rectangle with the new stimulating color, (c) is the complementary afterimage using the color pairs (red–cyan, green–magenta, and blue–yellow), and (d) is the afterimage predicted by our model.

Rather than reporting on experiments conducted in a small population of subjects, Fig. 3~6 can be alternatively used as the static UI for our tests, allowing the reader to serve as subject as well as judge. Readers can look at the subfigure (a) for 20s and then move to the subfigure (b), the afterimage shall appear in the reader's vision. Then readers can choose one image between subfigures (c) and (d) and pick the one that is more closely match with the afterimage in reader's vision. Readers shall enlarge the each figure to the full size of computer screen so that the interplay of different subfigures could be reduced to the minimum.

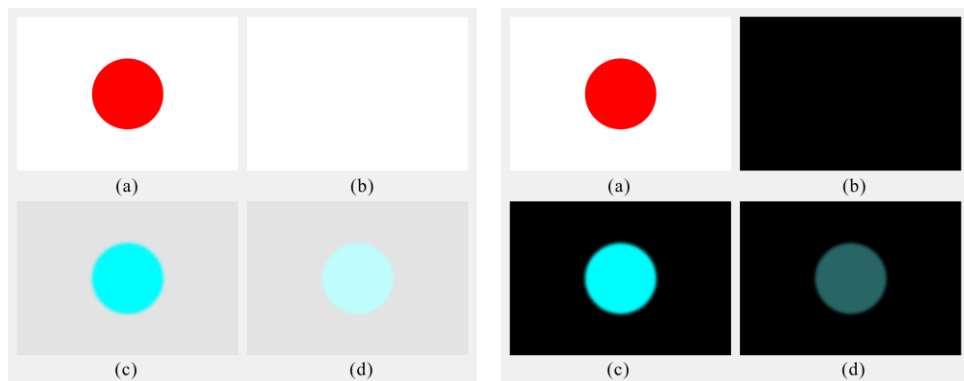

Fig. 3. Afterimage tests of a red circle over the white surround when the new stimulating color is white (left) and black (right).

Fig. 3 left shows an example of afterimages where both the old stimulating color $C_{OI}$ in the inducing field in (a) and new stimulating color $C_N$ in (b) are white, this is the case previous complementary afterimage theories most concern with. When the old stimulating color in the test field is red, the color of the test field in the afterimage (d) predicted by our model is $C_{AT}(0.76,1,1)$, which is slightly brighter than that $C_{CT}(0,0.9,0.9)$ in the complementary afterimage (c), where the subscript C stands for complementary.

When the new stimulating color is black (this case amounts to no light coming into eyes), as shown in Fig. 3 right, we are still able to see an after image of the circle with a darker green-blue color, $C_{AT}(0.16,0.4,0.4)$ as predicted by our model in (d). While the green-blue afterimage obtained with the complementary afterimage theory differs dramatically in strength from the afterimage in our vision.

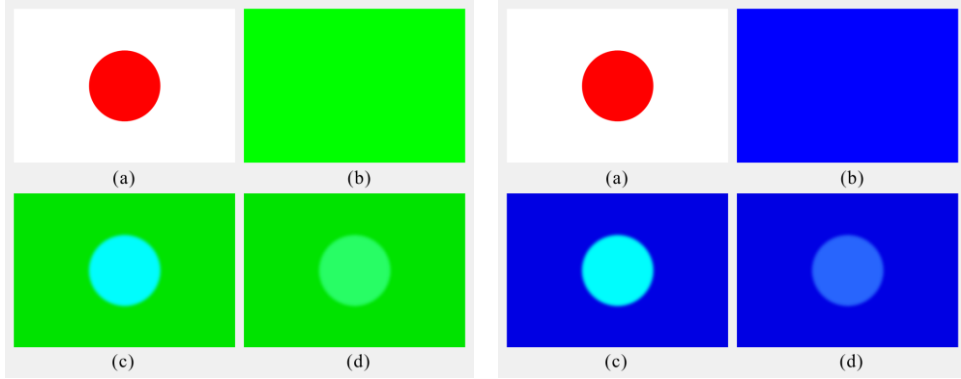

Fig. 4. Afterimage tests of a red circle over the white surround when the new stimulating color is green (left) and blue (right).

In Fig. 4 left, the new stimulating color is green, the afterimage color in the test field predicted by our model is $C_{AT}(0.16,1,0.4)$ in which the green component is dominant. Although it differs greatly from that $C_{CT}(0,0.9,0.9)$ in the complementary afterimage (c), it is more consistent with the afterimage appearing in our vision. In Fig. 4 right, the new stimulating color is blue, the color in the test field predicted by our model is $C_{AT}(0.16,0.4,1)$, which is also better than its counterpart in the complementary afterimage in (c).

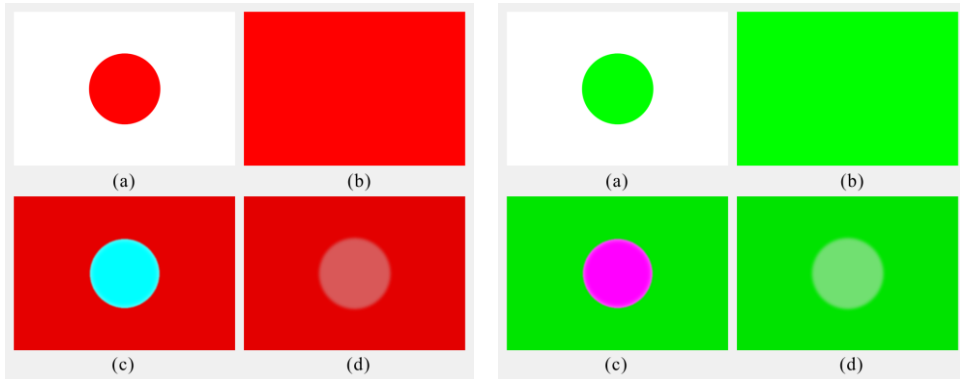

Fig. 5. Two afterimage tests in which the new stimulating color is the same with the old stimulating color in the test field in the original image.

Fig. 5 shows two examples where the inducing field is kept white, and the new stimulating color $C_N$ is the same with the old stimulating color $C_{OT}$ in the test field. In Fig. 5 left, both $C_N$ and $C_{OT}$ are red. With the parameter setting ($\alpha=0.6$, $\beta_T=0.35$) corresponding to this case, the color in the test field of the afterimage predicted our model is $C_{AT}(0.86,0.35,0.35)$ as shown in (d), which is more consistent with the afterimage in our vision compared with complementary afterimage shown in (c).

In Fig. 5 right, both $C_N$ and $C_{OT}$ are green, with the parameter setting ($\alpha=0.7$, $\beta_T=0.4$) corresponding to this case, the color in the test field of the afterimage predicted by our model is $C_{AT}(0.45,0.8875,0.45)$ as shown in (d), which is much better than its counterpart $C_{CT}(0.9,0,0.9)$ in the complementary afterimage shown in (c).

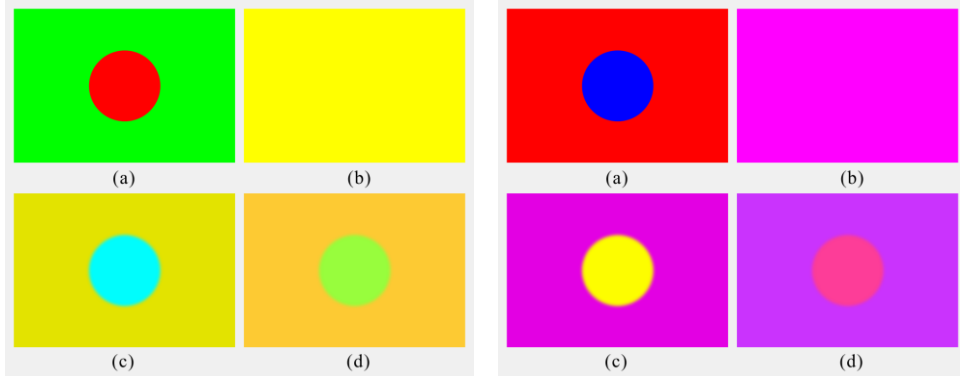

Fig. 6. Left: Afterimage test of a red circle over the green surround when the new stimulating color is yellow. Right: Afterimage test of a blue circle over the red surround when the new stimulating color is magenta.

In Fig. 6 we present two examples different from previous ones in that both old stimulating colors in the test and inducing fields and the new stimulating color are chromatic. With the colors settings $C_{OT}$=RED, $C_{OI}$=GREEN, and $C_N$=YELLOW, as shown in Fig. 6 left, the estimated colors in the test and inducing fields of the afterimage are $C_{AT}(0.76,1.0,0.4)$ and $C_{AI}(1.0,0.8,0.2)$, respectively, as shown in Fig. 6 left (d). With the colors settings $C_{OT}$=BLUE $C_{OI}$=RED, and $C_N$=MAGENTA(1,0,1), as shown in Fig. 6 right, the estimated colors in the test and inducing fields of the afterimage are $C_{AT}(1.0,0.4,0.76)$ and $C_{AI}(0.8,0.2,1.0)$, respectively, as shown in Fig. 6 right (d).

## 8. Discussions

Afterimages arise from interactions between old stimulating colors in the test and inducing fields at the simultaneous contrast stage and the interactions between colors in the previously-viewed field and currently-viewed field at the successive contrast stage, thus they are functions of old stimulating colors in the test and inducing fields as well as the new stimulating color, as illustrated by the formulae in (3) and (4). We should note that our current study focuses primarily on the simulation of the initial impression of afterimages rather than the dynamics of the fade out of afterimages. Since the mechanism of afterimages in our vision is not yet understood fully, this study is an attempt to numerically estimate afterimage colors, which can provide useful insights to what influence afterimage colors most.